\documentclass[10pt,twocolumn,letterpaper]{article}

\usepackage{iccv}
\usepackage{times}
\usepackage{epsfig}
\usepackage{graphicx}
\usepackage[dvipsnames]{xcolor}
\usepackage{amsmath}
\usepackage{amssymb}
\usepackage{stmaryrd}
\usepackage{tabularx,booktabs}
\usepackage{multirow}
\usepackage{lipsum}
\usepackage{array}
\usepackage{color, colortbl}
\usepackage{pifont}
\usepackage{flushend}
\usepackage{makecell}

\usepackage{sidecap}

\usepackage[accsupp]{axessibility} 

\usepackage[pagebackref=true,breaklinks=true,colorlinks,citecolor=blue,bookmarks=false]{hyperref}

\iccvfinalcopy 

\def\iccvPaperID{9024} 

\ificcvfinal\fi

\newcommand{\mcl}[1]{\ensuremath{\mathcal{#1}}}

\newcommand{\vect}[1]{\ensuremath{\textbf{#1}}}

\newcolumntype{C}{>{\centering\arraybackslash}X}
\newcolumntype{L}{>{\centering\arraybackslash}p{0.2\textwidth}}

\begin{document}

\title{Multi-View Radar Semantic Segmentation}

\author{Arthur Ouaknine$^{1,2}$

\and
Alasdair Newson$^1$

\and
Patrick Pérez$^2$

\and
Florence Tupin$^1$

\and
Julien Rebut$^2$\\

\\
$^1$LTCI, Télécom Paris, Institut Polytechnique de Paris, Palaiseau, France\\
$^2$valeo.ai, Paris, France\\

{\tt\small arthur.ouaknine@telecom-paris.fr}
}

\maketitle
\ificcvfinal\thispagestyle{empty}\fi

\begin{abstract}
    Understanding the scene around the ego-vehicle is key to assisted and autonomous driving.
    Nowadays, this is mostly conducted using cameras and laser scanners, despite their 
    reduced performance in adverse weather conditions.
    Automotive radars are low-cost active sensors that measure properties of surrounding objects, including their relative speed, and have the key advantage of not being impacted by rain, snow or fog. However, they are seldom used for scene understanding due to the size and complexity of radar raw data and the lack of annotated datasets.  
    Fortunately, recent open-sourced datasets have opened up research on classification, object detection and semantic segmentation with raw radar signals using end-to-end trainable models.
    In this work, we propose several novel architectures, and their associated losses, which analyse multiple ``views'' of the range-angle-Doppler radar tensor 
    to segment it semantically. 
    Experiments conducted on the recent CARRADA dataset demonstrate that our best model outperforms alternative models, derived either from the semantic segmentation of natural images or from radar scene understanding, while requiring significantly fewer parameters.
    Both our code and trained models are available at \url{https://github.com/valeoai/MVRSS}.
\end{abstract}

\begin{figure}[t]
   \includegraphics[width=1.0\linewidth]{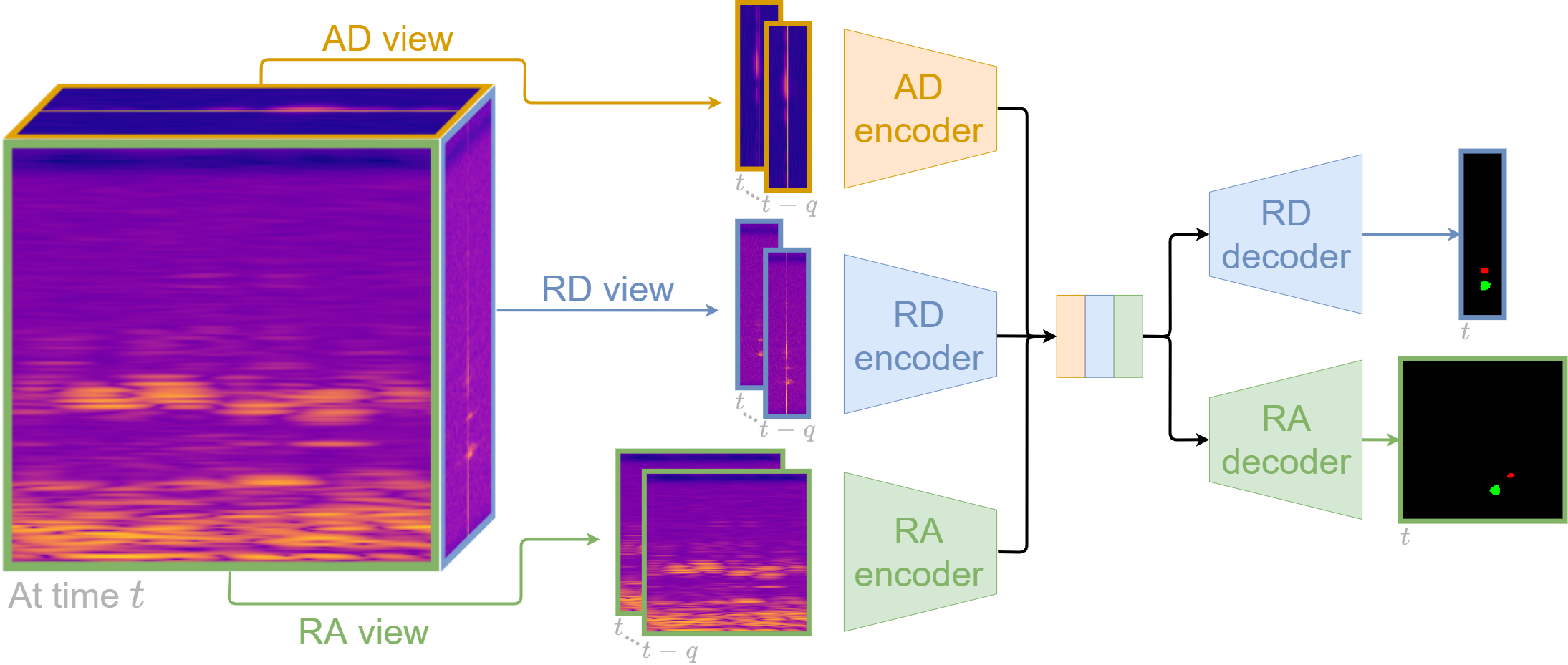}
   \caption{
   \textbf{Overview of our multi-view approach to semantic segmentation of radar signal}. At a given instant, radar signals take the form of a range-angle-Doppler (RAD) tensor. Sequences of $q+1$
   2D views of this data cube are formed and mapped to a common latent space by the proposed multi-view architectures. Two heads with distinct decoders produce a semantic segmentation of the range-angle (RA) and range-Doppler (RD) views respectively (`background' in black, `pedestrian' in red and `cyclist' in green in this example). 
   }
\label{fig:teaser}
\end{figure}

\section{Introduction}

Radar sensors have been used in the automotive industry for the last two decades, e.g., for automatic cruise control or blind spot detection. 
They have become the sensor of choice for applications requiring time to collision as they provide, besides localization, the relative velocity thanks to the Doppler information. However, these radars have been hindered previously by their poor angular resolution.

The shift from driving assistance to automated driving drastically increased the necessary level of performance, robustness and safety. 
Safety requires redundancy mechanisms at all levels of the system: From the sensing parts to the final decision modules. Redundancy at the sensor level can be reached using three sensors of different nature such as camera, lidar and radar.

While deep learning has brought about major progress in the automotive use of cameras and lidars -- for object detection and segmentation in particular -- it is only recently that it has also embraced radar signals. In fact, even though the radar technology has greatly improved, the signal processing pipeline has remained the same for years. 
This sensor is now a source of interest since public datasets have been released \cite{barnes_oxford_2020, caesar_nuscenes_2020, gao_experiments_2019, meyer_automotive_2019, nowruzi_deep_2020, ouaknine_carrada_2020, sheeny_radiate_2020}.

Radar scene understanding is in its infancy, but it provides key information to compensate for weaknesses of the other sensors. The radar data presented as a range-angle-Doppler (RAD) tensor contains signatures of the objects surrounding the car with enough details to distinguish them in the representation. 
Unlike object detection using bounding boxes, semantic segmentation is appropriate for this task, since the object signatures have extremely variable sizes and may be mixed up due to the sensor's resolution.

In this work, we propose an approach to multi-view radar semantic segmentation, illustrated in Fig.\,\ref{fig:teaser}, that exploits the entire data  while addressing the challenges of its large volume and high level of noise. 
The segmentation is performed on the range-Doppler and range-angle views, which suffices to deduce the localisation and the relative speed of objects.
Our first contribution is a set of lightweight neural network architectures designed for multi-view semantic segmentation of radar signal. The second contribution is a set of loss terms to train models on these tasks while preserving coherence between the multi-view predictions. 
Experiments show that our best model outperforms other methods considered for radar semantic segmentation in both quantitative and qualitative evaluations.

We present automotive radar sensing and related works in \S\,\ref{sec:background}, then our contributions and the methods that we compare (\S\,\ref{sec:method_archis}), experiments (\S\,\ref{sec:experiments}) and conclusions (\S\,\ref{sec:conclusions}).

\section{Background}
\label{sec:background}

\subsection{Automotive radar sensing}
\label{sec:radar_theory}

A radar is an \textit{active} sensor that emits electromagnetic waves, which are received back after being reflected in the environment \cite{brooker_understanding_2005, ghaleb_micro-doppler_2009}.
Standard in automotive, a frequency-modulated-continuous-wave (FMCW) radar emits a sequence of frequency-modulated signals called chirps.
The received signal in the time domain is recorded in a 3D tensor indicating the chirp index, the chirp sampling and the corresponding receiver antenna index. It is usually named analog-to-digital converter (ADC) signal.
The object distance is extracted using a Fourier transform along the chirp sequence (Range-FFT). 
A Doppler-FFT is then applied along the chirp sampling axis to estimate the phase difference and deduce the radial velocity of the reflective surface. 
Finally, an Angle-FFT processes the signal through the pairs of antennas to estimate the view angle to the object.
This sequence of FFTs results in the range-angle-Doppler (RAD) tensor, a 3D data cube of complex numbers where each axis amounts to discretised values of the corresponding physical measurement.
With conventional FMCW radars, the RAD tensor is usually not available as it is too computing intensive to estimate. CFAR algorithm \cite{rohling_radar_1983} is typically applied to extract objects in range-Doppler while a sparse point cloud is obtained using beamforming.

\smallskip\noindent\textbf{From RAD tensor to multiple views.~} 
The intensities (squared modulus) in the RAD tensor provide information on the power backscattered by moving objects
; See a visualization in Appendix \ref{sec:rad_tensor_vis}.
Due to the coherent imaging, the backscattered signal presents strong fluctuations also known as speckle phenomenon and well studied by Goodman \cite{goodman_fundamental_1976}. This phenomenon can be modeled as a multiplicative noise. The averaging of few samples (operation known as multi-looking) reduces the noise strength, whereas the logarithmic transform induces a variance stabilization of the resulting signal \cite{deledalle_mulog_2017}.

Let $\vect{X}^{\text{RAD}}$ be the complex RAD tensor.
Averaging its intensities across one of its dimensions leads to three possible 2D views: RA, RD and AD (Fig.\,\ref{fig:teaser}). 
For instance, the RA view, expressed in decibels, is defined as:
\begin{equation}
    \vect{x}^{\text{RA}}[r,a] = 10 \log_{10}
    \Big(
    \frac{1}{N_{\text{D}}} \sum_{d=1}^{N_{\text{D}}} 
    \big |\vect{X}^\text{RAD}[r,a,d] \big|^2
    \Big),
\end{equation}
with $|.|$ the modulus and $N_{\text{D}}$ the number of Doppler bins. 
Turning RAD tensors into views aims both to compress substantially the data representation and to reduce its noise. In practice, this reduces the size of the data by a factor of 50.

\subsection{Related work}
Low-cost FMCW radar is used in applications such as hand-gesture recognition \cite{dekker_gesture_2017, hazra_radar_2019, kim_hand_2016, lei_continuous_2020, scherer_tinyradarnn_2020, sun_automatic_2019,  wang_interacting_2016, wang_rammar_2019, zhang_u-deephand_2019, zhang_latern_2018}, human activity recognition \cite{zhu_classification_2020}, fall motion recognition \cite{shankar_radar-based_2019}, in-vehicle passenger detection \cite{in-vehicle-radar} or surveillance \cite{radar-Traffic-Monitoring}. This section will focus on automotive applications.

Raw radar signal is rich but complicated
and noisy. Its processing pipeline can deliver a sparse point cloud (PC). Scene analysis can be done at various stages of this chain.

\smallskip\noindent\textbf{Point-cloud architectures.~}
Point clouds are the most common interfaces of commercialized automotive radar. 
This representation has been broadly explored for object detection \cite{danzer_2d_2019, meyer_deep_2019}, non-line-of-sight occluded object localisation and tracking \cite{scheiner_seeing_2020} and occupancy grid segmentation in the bird's eye view (BEV) \cite{lombacher_semantic_2017, prophet_semantic_2020,  prophet_semantic_2019, schumann_scene_2020, sless_road_2019}.
Despite good performances, PC approaches suffer from the radar pre-processing inefficiency.
Filtering is usually performed to remove noise and simplify the signal representation. Unfortunately, this leads to the loss of useful information. 

\smallskip\noindent\textbf{Single-view methods.~}
An alternative to PC representation is to exploit the signal before CFAR filtering. Since the 3D RAD representation is cumbersome, 
it is usually considered by slicing or aggregating the tensor along one dimension.

The range-angle (RA) view, with its intuitive polar representation of the scene, is preferable for tasks like 
object classification \cite{patel_deep_2019}, detection \cite{dong_probabilistic_2020} and localisation \cite{wang_rodnet_2021}, and for odometry \cite{aldera_what_2019}. In particular, \cite{kaul_rss-net_2020} proposes a method to segment static and moving objects in RA view, which we compare to (see Section \ref{sec:radar_based_methods} for details).

The range-Doppler (RD) view is particularly useful since it indicates the relative speed of scene reflectors at each range bin. It is not widely explored though because of its less direct interpretability.  
Recent automotive works use Doppler spectrograms for vehicle classification \cite{capobianco_vehicle_2018} and RD views for object detection \cite{ng_range-doppler_2020, zhang_object_2020} or source separation \cite{changbo_radar_2020}. Brodeski \etal \cite{brodeski_deep_2019} also propose a two-stage method to detect objects in RD view and infer their position in simple and limited contexts.

Single-view methods suffer from a lack of information, as they do not exploit the entire radar data. Moreover, predictions on a single view are not able to provide both the position and the relative velocity of a detected object. 

\smallskip\noindent\textbf{Multi-view methods.~}
Recent works exploit the entire information in the RAD tensor.
Handling this bulky representation is challenging.
Palffy \etal \cite{palffy_cnn_2020} use a multi-stage method on only a sub-part of the tensor and detect objects in camera images. Major \etal \cite{major_vehicle_2019} create 2D views by sum-pooling over each of the tensor's axes. Their multi-view representations are processed by a single network specialised in RA object detection.
In the same vein, Gao \etal \cite{gao_ramp-cnn_2020} also pre-process the RAD tensor into summed 2D views. 
Then, specialised auto-encoders extract features from each view which are
fused to localise objects in the RA view. See further details on this approach 
in Sec.\,\ref{sec:radar_based_methods}.

To the best of our knowledge, there is no previous work on multi-view radar semantic segmentation. 
In the next sections, we will describe our method which is able to both process the entire RAD tensor and segment radar multi-views 
to predict localisation and relative velocity of objects.

\section{Methods and architectures}
\label{sec:method_archis}

We now present in Sections\,\ref{sec:img_based_methods}  and \ref{sec:radar_based_methods} several methods for image segmentation and radar scene understanding, to which we compare our work. They are chosen for their performance and their relevance to the radar semantic segmentation task.
Except RSS-Net \cite{kaul_rss-net_2020}, 
these architectures were not originally designed for radar semantic segmentation, nor to handle multiple views of a data volume.
Consequently, for each of them, we have trained two models independently for range-Doppler and range-angle segmentation respectively. More details are provided in Section \ref{sec:method_modif}.

In Section\,\ref{sec:archi_ours}, we then introduce the three proposed architectures for multi-view radar semantic segmentation.

\subsection{Image-based competing methods}
\label{sec:img_based_methods}

Long \etal \cite{long_fully_2015} propose Fully Convolutional Networks (FCN), consisting of convolutional and down-sampling layers followed by transposed convolutions (``up-convolutions''). The final representations are processed by a 1D convolution with softmax to predict a category for each pixel. 
Several versions are proposed 
depending on the feature-map scales used to generate the output.
FCN has been used for semantic segmentation of radar data in \cite{ouaknine_carrada_2020}, where FCN-8s version achieves the best performance.

The U-Net architecture \cite{ronneberger_u-net_2015} is composed of equal-depth down-sampling and up-sampling pathways linked by skip connections.
Originally used for medical images, it is especially well suited for small-object segmentation.

The DeepLabv3+ 
\cite{chen_encoder-decoder_2018} is a popular encoder-decoder model for semantic segmentation of natural images. 
The encoder uses ``atrous'' separable convolutions 
which increase the receptive field of the network. The proposed Atrous Spatial Pyramidal Pooling (ASPP) layer \cite{chen_deeplab_2017} combines atrous convolutions with various dilation rates to learn multi-scale features followed by a 1D convolution.

\subsection{Radar-based competing methods}
\label{sec:radar_based_methods}

Kaul \etal \cite{kaul_rss-net_2020} propose RSS-Net, specialised in radar semantic segmentation, in particular for range-angle representations. Its architecture is similar to DeepLabv3+ with an encoder composed of 8 atrous convolutional layers, an ASPP module and a convolutional decoder with up-sampling.
The architecture is designed
to reduce the dimension of the feature maps in the encoder, leading to excellent performance for range-angle BEV semantic segmentation. 

Gao \etal \cite{gao_ramp-cnn_2020} propose the Radar Multiple-Perspective Neural Network (RAMP-CNN) for object detection in RA representations. They aggregate the RAD tensor into 2D radar views which are processed by separate encoder-decoders with 3D convolutional layers.
The final range-angle features are processed by a 3D inception module. RAMP-CNN achieves state-of-the-art performance in localising and classifying multiple objects in range-angle views.

\subsection{Proposed multi-view methods}
\label{sec:archi_ours}

\begin{figure*}
\begin{center}
\includegraphics[width=0.90\linewidth]{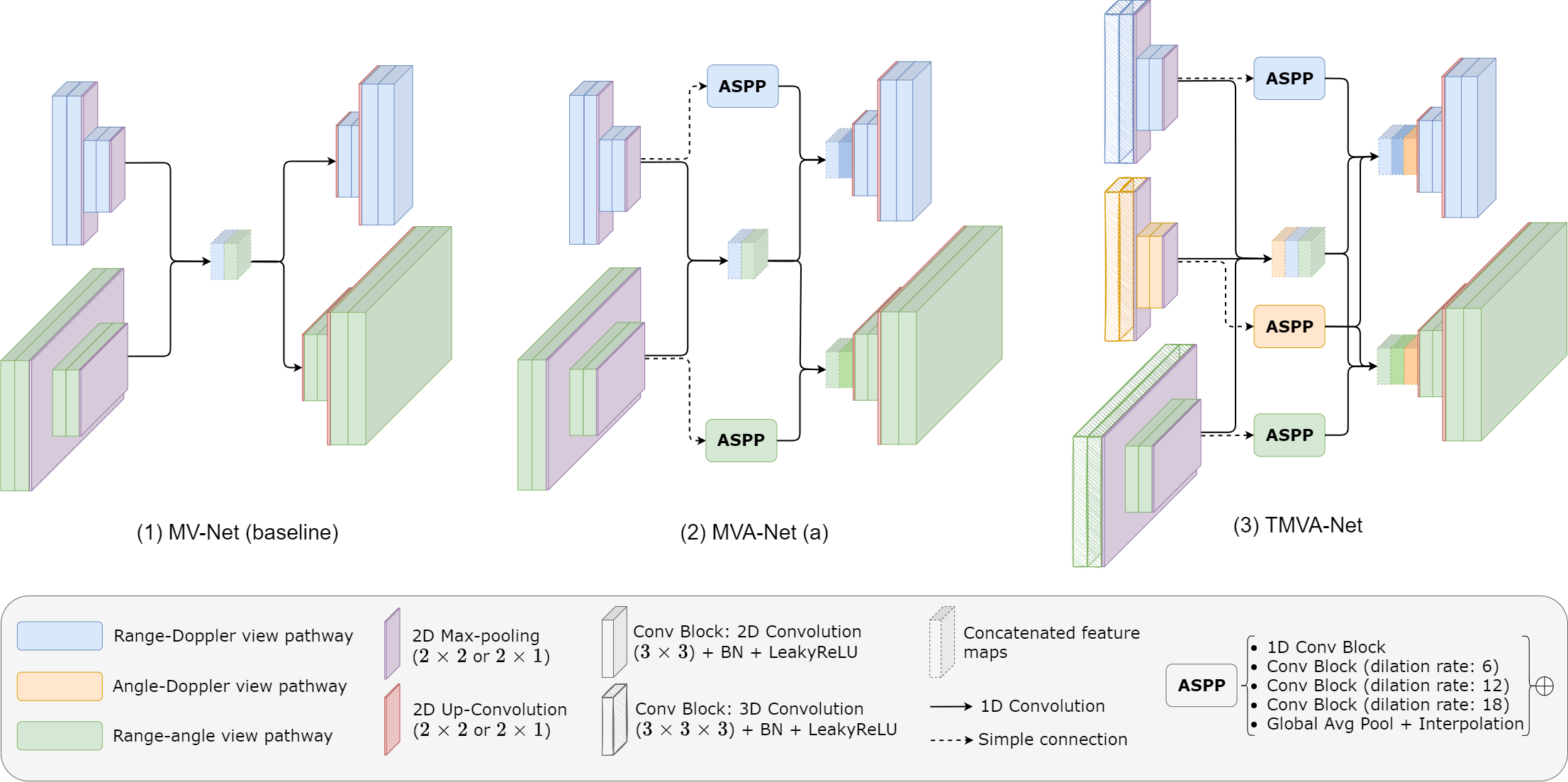}
    \caption{\textbf{Multi-view architectures for radar semantic segmentation.} 
    (1) The multi-view network (MV-Net), considered as a baseline, is composed of two encoders, two decoders and a common latent space. 
    (2) The MVA-Net(a) has an additional ASPP module for each view pathway; See text for variant MVA-Net(b); 
    (3) The TMVA-Net architecture is similar to the MVA-Net(a) with an additional angle-Doppler encoder while exploiting the temporal dimension with 3D convolutions at the top of the encoders.
    The detailed architectures are provided in Appendix \ref{sec:archi_details}.
    }
\label{fig:architectures}
\end{center}
\end{figure*}

We propose three lightweight neural network architectures specialised in multi-view radar semantic segmentation, whose general principle is illustrated in Fig.\,\ref{fig:teaser}. 
They take a temporal stack of radar views as their input and process them with dedicated encoders.
The generated feature maps are fused in a shared latent space from which different decoders predict semantic segmentation maps for each output view.
Since neither encoders nor decoders share weights,
we keep them simple. 
Thus, there are specialised parts of the network in each view, and a reasonable number of parameters altogether. Further details about the architectures are provided in Appendix \ref{sec:archi_details}.

\smallskip\noindent\textbf{Multi-view network (MV-Net).~}  We first propose (see Figure \ref{fig:architectures}\,(1)) a baseline in the form of a double encoder-decoder architecture that processes stacked RD and RA views and predicts simultaneously the RD and RA semantic segmentation maps. It is important to note that we can deduce the third view (AD) from the RA and RD views, and it is therefore not necessary to segment it in the output. This is the case for all the architectures we propose.
 
Each encoder is composed of two blocks, each one with two sequences of convolution, batch normalisation and activation layers. The two blocks are separated by a max-pooling operation to down-sample the feature maps along the range axis (Doppler's resolution being lower, we keep it unchanged).
The feature maps from both encoders are down-sampled, processed by a 1D convolution and stacked into a common latent space. Since the input views are stacked according to the time dimension, this linear combination of the feature maps aims to learn temporal correlations.
The features in the shared latent space are then processed with 1D convolution layers and used as the input of each decoder.
There are two decoders predicting respectively the range-Doppler and range-angle semantic segmentation maps. Each one is composed of two blocks with two sequences of convolution, batch normalisation and activation layers. The up-sampling between the blocks is carried out by up-convolutions. A final 1D convolution performs a combination of the outputs of each decoder to generate $K$ feature maps, where $K$ is the number of classes. A softmax operation is then applied pixel-wise to the $K$ feature maps generating soft masks.

\smallskip\noindent\textbf{Multi-view network with ASPP modules (MVA-Net).~} 
The ASSP module \cite{chen_deeplab_2017} used in DeepLabv3 \cite{chen_encoder-decoder_2018} allows features to be jointly learned at different scales at a given depth of the network with no need for larger kernels or additional parameters. As shown in RSS-Net \cite{kaul_rss-net_2020}, it is well suited for radar semantic segmentation since the objects' signatures can vary considerably.
Our second architecture, MVA-Net, introduces the use of this ASSP module at the end of each decoder of our MV-Net baseline.  
The generated multi-scale feature maps are concatenated, processed by a 1D convolution
and stacked to the input of each corresponding decoder.
In effect, we propose two variants of MVA-Net: MVA-Net(a), shown in Figure \ref{fig:architectures}\,(2), consists in a double encoder-decoder architecture with ASPP modules to process and segment RD and RA views; MVA-Net(b) has an additional encoding branch learning features from the AD view. 
Similarly to the other encoding branches, the AD backbone generates features that are stacked in the common latent space. However, the outputs of its ASPP module contain both the angle and Doppler information. Thus the multi-scale feature maps from the AD pathway are stacked to the inputs of both the RD and RA decoders.

\smallskip\noindent\textbf{Temporal multi-view network with ASPP modules (TMVA-Net).~} 
The temporal dimension provides valuable information for radar semantic segmentation. It helps in estimating the shape of an object's signature despite high noise levels, and distinguishing objects that are close together with similar velocities.
Our third architecture, which we will show to have the best performances, TMVA-Net in Fig.\,\ref{fig:architectures}\,(3), extends MVA-Net by explicitly leveraging the temporal dimension. 
The first block's 2D convolutions are replaced by 3D convolutions in each encoder branch, making it able to learn the spatio-temporal characteristics with limited increase in the number of parameters. Since 3D convolutions require a large number of parameters, full 3D-convolutional encoders, such as in \cite{gao_ramp-cnn_2020}, have not been retained.
Hence, TMVA-Net is composed of three encoders with 3D and 2D convolutions, one for each input view.
Each one of them has a dedicated ASPP module. The feature maps generated from each encoding backbone are stacked into a shared latent space. From there, two decoders segment respectively the RD and RA views. They take as input the stacked features from the processed latent space and the multi-scale feature maps from the dedicated ASPP modules.

\subsection{Losses}
\label{losses}

In what follows, we use the generic notation $f_\theta(\vect{x})\!=\!\vect{p}$ for a segmentation model with parameters $\theta$, input $\vect x$ and output $\vect p$. Training $f_{\theta}$ amounts to minimising w.r.t. $\theta$ a suitable loss function, given training examples $\vect x$ with ground truth $\vect y$.  
The architectures presented in Secs.\,\ref{sec:img_based_methods}-\ref{sec:radar_based_methods} take single-view inputs stacked in the temporal dimension and predict, for each target view, a soft segmentation mask with class ``probabilities" for each bin. 
For instance, the output of a model processing only the RA view is $f_\theta(\vect{x}^{\text{RA}})\!=\! \vect{p}^{\text{RA}}\!\in\![0,1]^{M\times N\times K}$, if $M\!\times\!N$ is the size of the view and $K$ the number of classes.
Our architectures, detailed in Sec.\,\ref{sec:archi_ours}, take instead multi-view inputs: either $\vect{x}\!=\!(\vect{x}^{\text{RD}}, \vect{x}^{\text{RA}})$ or $\vect{x}\!=\!(\vect{x}^{\text{RD}}, \vect{x}^{\text{AD}}, \vect{x}^{\text{RA}})$. 
In both cases, their goal is to predict soft masks $\vect{p}\!=\!(\vect{p}^{\text{RD}}, \vect{p}^{\text{RA}})$ for both RD and RA views.

The following section details the loss functions applied to each segmented view to train the proposed architectures. We also introduce a `coherence' loss to enforce consistency between the predictions over the two views of the scene. 
Finally a combination of these loss terms is proposed.

\smallskip\noindent\textbf{Weighted Cross Entropy.~}
\label{sec:wce}
Semantic segmentation models that predict a score for each class at each pixel are usually trained by minimising a Cross-Entropy (CE) loss function.
This loss is not ideal for unbalanced segmentation tasks such as the radar semantic segmentation, since the optimisation process tends to focus on the classes that are most represented. 
In the present case, background and speckle noise dominate, in comparison to the signatures of the objects we wish to detect.
In the same manner as RSS-Net \cite{kaul_rss-net_2020}, we employ a weighted Cross-Entropy (wCE) loss to tackle this issue.

Given a training example $\vect x$, let $\vect{y}\in \{0,1\}^{M \times N \times K}$ be its one-hot ground truth and $f_\theta(\vect{x}) = \vect{p} \in [0,1]^{M \times N \times K}$ the associated prediction. 
The wCE loss function is defined as:
\begin{equation}
\mcl{L}_{\text{wCE}} (\vect{y}, \vect{p}) = - \frac{1}{K} \sum_{k=1}^{K} w_k \!\!\!\!\sum_{(m,n)\in\Omega}\!\!\!\!\vect{y}[m,n,k] \log \vect{p}[m,n,k],
\label{eq:lossWCE}
\end{equation}
where $\Omega = \llbracket 1,M\rrbracket \times \llbracket 1,N\rrbracket$,
and $w_k$'s are normalized positive weights. Weight $w_k$ is inversely proportional to the frequency of class $k$ in the training set, that is $w_k\propto \big(\sum_{\vect y}\sum_{(m,n)\in\Omega}\vect{y}[m,n,k]\big)^{-1}$. The fewer the bins with ground-truth class $k$, the larger $w_k$ becomes. 

\smallskip\noindent\textbf{Soft Dice.~}
\label{sec:sdice}
Object signatures in radar representations often correspond to small regions. This is a well known issue in medical image segmentation, where the Dice metric (detailed in Section \ref{sec:data_metrics}) is usually reformulated in a function called Dice loss, ranging between 0 and 1. 
Milletari \etal \cite{milletari_v-net_2016} have proposed the Soft Dice (SDice) loss defined as:
\begin{equation}
\mcl{L}_{\text{SDice}} = \frac{1}{K} \sum_{k=1}^{K} \Bigg[ 1 - \frac{2 \sum_{(m,n)} \vect{y}[m,n,k] \vect{p}[m,n,k]}{\sum_{(m,n)} \vect{y}^2[m,n,k] + \vect{p}^2[m,n,k]} \Bigg],
\end{equation}
where $(m,n) \in \Omega$, 
as in Eq.\,\ref{eq:lossWCE}. 
This formulation has proved useful for 2D and 3D medical image semantic segmentation, including for small objects. 

\smallskip\noindent\textbf{Coherence.~}
\label{sec:col}
The objective of multi-view radar semantic segmentation is to simultaneously segment several views of the aggregated RAD tensor. 
The objects we wish to detect are observed in the different radar views, thus it is clear that a certain coherence must be maintained between the segmented views. For example, one view should not represent a pedestrian, while another represents a cyclist. We introduce a coherence loss (CoL) to preserve a consistency between the predictions of the model. The procedure to calculate this loss is illustrated in Appendix \ref{sec:col_figure}.

Let $(\vect{p}^{\text{RD}}, \vect{p}^{\text{RA}})$ be the segmentation maps predicted by the model $f_\theta$ after the softmax operation. These two maps are aggregated by applying a $\max(.)$ operator along the axis that they do not share (either the Doppler or the angle). The two resulting maps of same size, denoted  $\tilde{\vect{p}}^{\text{RD}}$ and $\tilde{\vect{p}}^{\text{RA}}$, contain the highest probability of each range bin for each class. In other words, they indicate if the model predicts a high probability to observe a category at a given distance. The coherence loss is the mean squared error (MSE) between these maximum range probability vectors. It encourages the network to predict high probability values at the same distance and in the same class for both views. The CoL, in the interval $[0,1]$, is defined as:
\begin{equation}
\mcl{L}_{\text{CoL}}(\vect{p}^{\text{RD}}, \vect{p}^{\text{RA}}) = \big \lVert \tilde{\vect{p}}^{\text{RD}} -
\tilde{\vect{p}}^{\text{RA}}\big \rVert^2_{\mathrm{F}},
\end{equation}
where $\|\cdot\|_{\mathrm{F}}$ denotes the Frobenius norm.

\smallskip\noindent\textbf{Combination of losses.~}
\label{sec:combi_loss}
The CE loss is specialised in pixel-wise classification and does not consider spatial correlations between the predictions. The SDice is particularly effective for shape segmentation, but it is difficult to optimise as a single loss function due to its gradient formulation.
Finally, the CoL is useful where neither the CE nor the SDice is able to leverage a coherence between the prediction of the RD and the RA views. To combine the different strengths of these losses, we propose the following final loss to train multi-view architectures:
\begin{equation}
\mcl{L} = \lambda_{\text{wCE}} (\mcl{L}_{\text{wCE}}^{\text{RD}} + \mcl{L}_{\text{wCE}}^{\text{RA}}) + \lambda_{\text{SDice}} (\mcl{L}_{\text{SDice}}^{\text{RD}} + \mcl{L}_{\text{SDice}}^{\text{RA}}) + \lambda_{\text{CoL}} \mcl{L}_{\text{CoL}},
\end{equation}
where $\lambda_{\text{wCE}}$, $\lambda_{\text{SDice}}$ and $\lambda_{\text{CoL}}$ are weighting factors set empirically.

\section{Experiments}
\label{sec:experiments}

We present in this section the experimental evaluation of our models.
We describe first the datasets and the evaluation metrics that we utilize. 
We then explain the modifications made to the methods we compare to. 
Finally, we give details concerning the experiments and analyse their results quantitatively and qualitatively. 

\subsection{Datasets and evaluation metrics}
\label{sec:data_metrics}

\smallskip\noindent\textbf{Datasets.~} 
The CARRADA dataset \cite{ouaknine_carrada_2020} contains synchronised camera and automotive radar recordings with 30 sequences of various scenarios with one or two moving objects. The radar views are annotated using a semi-automatic pipeline. 
This is the only publicly-available dataset providing RAD tensors and dense semantic segmentation annotation for both RD and RA views. The objects are separated into four categories: \textit{pedestrian}, \textit{cyclist},  \textit{car} and \textit{background}. 
The provided RAD tensors have dimensions $256 \times 256 \times 64$.
The experiments presented in Section \ref{sec:train_and_res} use the dataset splits proposed by the authors, denoted CARRADA-Train, CARRADA-Val and CARRADA-Test.

For qualitative evaluation only, we also employ in-house sequences of more complex urban scenes with synchronised camera and radar data. For these sequences, the RAD tensors have the same dimensions as in  CARRADA while the resolution in range is divided by two. The radar views are not annotated, which does not allow quantitative evaluation.

\smallskip\noindent\textbf{Evaluation metrics.~} 
A classic performance metric in semantic segmentation is the intersection over union (IoU): Given annotated test inputs, the IoU for a given class is defined as the percentage $\frac{|A \cap B|}{|A \cup B|}$, where $A$ is the set of locations predicted as stemming from this class and $B$ is the ground-truth set of locations for the same class. From the perspective of a single object in a given scene, the IoU measures how well and how completely it is segmented. 
Averaging this metric over all classes yields the mean IoU (mIoU) score.
Another related, yet slightly different metric, is provided by the Dice score: For a given class and with same notations as above, it is defined as $\frac{2|A \cap B|}{|A| + |B|}$. For global performance, it is averaged over all classes into the mean Dice (mDice). Seeing segmentation as a local 1-\textit{vs}.-all classification problem for each class, the Dice amounts to the harmonic mean of the precision and recall (a.k.a. F1 score).
The IoU and Dice metrics are considered as complementary; 
We report both of them in our experiments.

\begin{figure}[t]
\centering
\includegraphics[width=0.98\linewidth]{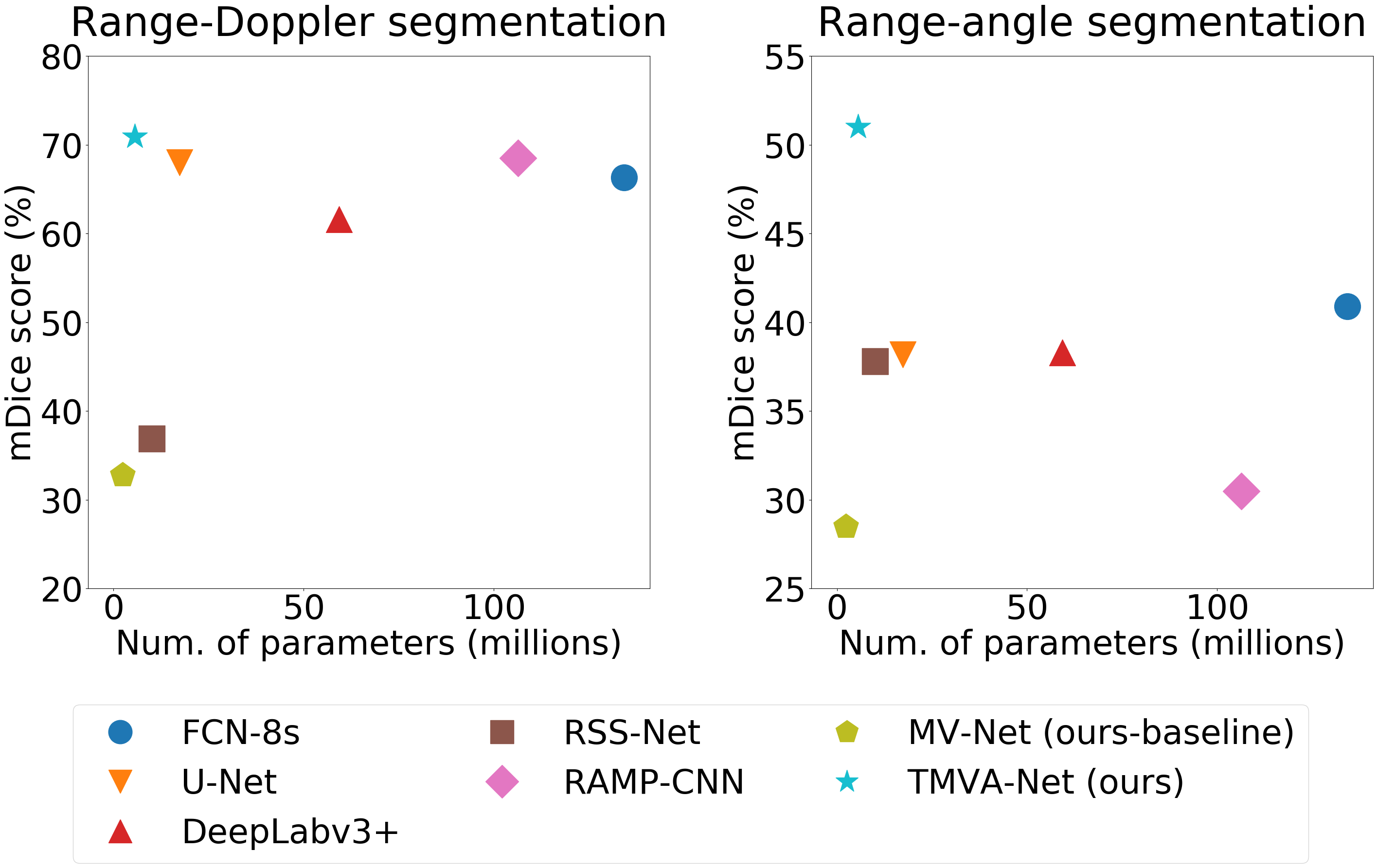}
    \caption{\textbf{Performance-\textit{vs}.-complexity plots for all methods in RD and RA tasks.} Performance is assessed by mDice (\%) and complexity by the number of parameters (in millions) \textit{for a single task}. 
    Top-left models are the best performing and the lightest. 
    Only our models, MV-Net and TMVA-Net, are able to segment both views simultaneously. For all the other methods, two distinct models must be trained to address both tasks, which doubles the number of actual parameters.}
\label{fig:perf_params}
\end{figure}

\begin{table*}[t]
\def\arraystretch{1.1}
\setlength\tabcolsep{3pt}
\scriptsize
\begin{tabularx}{\textwidth}{c @{\hskip 0.1in} c @{} r @{\hskip 0.2in} CCCC|C| CCCC|C @{}}
\toprule
 \multirow{5}{*}{View} & \multirow{5}{*}{Method} & \multirow{5}{*}{\# Param. (M)} & \multicolumn{5}{c}{IoU (\%)} &  \multicolumn{5}{c}{Dice (\%)} \\
 \cmidrule(lr{3pt}){4-8}
 \cmidrule(l{3pt}r{3pt}){9-13}
 & & & \rotatebox{90}{Bkg.} & \rotatebox{90}{Ped.} & \rotatebox{90}{Cyclist} & \rotatebox{90}{Car} & \rotatebox{90}{\textbf{mIoU}} 
 & \rotatebox{90}{Back.} & \rotatebox{90}{Ped.} & \rotatebox{90}{Cyclist} & \rotatebox{90}{Car} & \rotatebox{90}{\textbf{mDice}} \\ 
\midrule
    \multirow{7}{*}{\textbf{RD}} 
    & FCN-8s \cite{long_fully_2015} &  134.3~~ & 99.7 & 47.7 & 18.7 & 52.9 &  54.7  &  99.8 & 24.8 & 16.5 & 26.9 &  66.3 \\
& U-Net \cite{ronneberger_u-net_2015} &  17.3~~  & 99.7 & \underline{\textcolor{blue}{51.0}} & \textbf{\textcolor{red}{33.4}} & 37.7   &  55.4   &   99.8 & \underline{\textcolor{blue}{67.5}} & \textbf{\textcolor{red}{50.0}} & 54.7  &  68.0 \\
    & DeepLabv3+ \cite{chen_encoder-decoder_2018} &  59.3~~  & 99.7 & 43.2 & 11.2 & 49.2   &  50.8   &   99.9 & 60.3 & 20.2 & 66.0  &  61.6 \\
    & RSS-Net &  10.1~~  &  99.3 & 0.1 & 4.1 & 25.0 &  32.1   &   99.7 & 0.2 & 7.9 & 40.0 &  36.9 \\
& RAMP-CNN &  106.4~~  &  99.7 & 48.8 & 23.2 & \textbf{\textcolor{red}{54.7}} &  \underline{\textcolor{blue}{56.6}}   &   99.9 & 65.6 & 37.7 & \textbf{\textcolor{red}{70.8}}  &  \underline{\textcolor{blue}{68.5}} \\
    & MV-Net (ours-baseline) &  2.4*  & 98.0 & 0.0 & 3.8 & 14.1  &  29.0  &   99.0 & 0.0 & 7.3 & 24.8  &  32.8 \\
& TMVA-Net (ours) &  5.6*  & 99.7 & \textbf{\textcolor{red}{52.6}} & \underline{\textcolor{blue}{29.0}} & \underline{\textcolor{blue}{53.4}} & \textbf{\textcolor{red}{58.7}} & 99.8 & \textbf{\textcolor{red}{68.9}} & \underline{\textcolor{blue}{45.0}} & \underline{\textcolor{blue}{69.6}} & \textbf{\textcolor{red}{70.9}} \\
    \midrule
\multirow{7}{*}{\textbf{RA}} 
    & FCN-8s \cite{long_fully_2015} &  134.3~~  &  99.8 & 14.8 & 0.0 & \underline{\textcolor{blue}{23.3}} &  \underline{\textcolor{blue}{34.5}}  &  99.9 & 25.8 & 0.0 & \underline{\textcolor{blue}{37.8}}  &  \underline{\textcolor{blue}{40.9}} \\
    & U-Net \cite{ronneberger_u-net_2015} &  17.3~~ & 99.8 & \underline{\textcolor{blue}{22.4}} & \textbf{\textcolor{red}{8.8}} & 0.0  &  32.8   &   99.9 & \underline{\textcolor{blue}{36.6}} & \textbf{\textcolor{red}{16.1}} & 0.0  &  38.2 \\
    & DeepLabv3+ \cite{chen_encoder-decoder_2018} &  59.3~~  & 99.9 & 3.4 & 5.9 & 21.8  &  32.7  &   99.9 & 6.5 & 11.1 & 35.7  &  38.3 \\
& RSS-Net &  10.1~~  & 99.5 & 7.3 & 5.6 & 15.8 &  32.1   &   99.8 & 13.7 & 10.5 & 27.4  &  37.8 \\
    & RAMP-CNN &  106.4~~  &  99.8 & 1.7 & 2.6 & 7.2   &  27.9   &   99.9 & 3.4 & 5.1 & 13.5  &  30.5 \\
    & MV-Net (ours-baseline) &  2.4*  & 99.8 & 0.1 & 1.1 & 6.2  &  26.8  & 99.0 & 0.0 & 7.3 & 24.8 &  28.5 \\
& TMVA-Net (ours) &  5.6*  & 99.8 & \textbf{\textcolor{red}{26.0}} & \underline{\textcolor{blue}{8.6}} & \textbf{\textcolor{red}{30.7}} & \textbf{\textcolor{red}{41.3}} & 99.9 & \textbf{\textcolor{red}{41.3}} & \underline{\textcolor{blue}{15.9}} & \textbf{\textcolor{red}{47.0}} & \textbf{\textcolor{red}{51.0}} \\
\bottomrule
\end{tabularx}
\vspace{2pt}
 \caption{\textbf{Semantic segmentation performance on the CARRADA-Test dataset for range-Doppler (RD) and range-angle (RA) views}.
 The number of trainable parameters (in millions) for each method corresponds to a single view-segmentation model; Two such models, one for each view, are required for all methods but ours. In contrast, the number of parameters reported for our methods (`*') corresponds to a single model that segments both RD and RA views.
 The RSS-Net and RAMP-CNN methods have been modified to be trained on both tasks (see Sec.\,\ref{sec:method_modif}). Performances are evaluated with the Intersection over Union (`IoU') and the Dice score per class, and their averages, `mIoU' and `mDice', over the four classes.
 The best scores are in red and bold type, the second best in blue and underlined. }
 \label{table:main_quanti_results}
\end{table*}

\subsection{Implementation of competing methods}
\label{sec:method_modif}

This section describes the architectures we use for comparisons. 
For each method, one model is trained specifically for single-view semantic segmentation of either RD or RA. Details concerning pre-processing procedures are provided in Appendix \ref{sec:preproc_procedures}.

The non-radar-based architectures have been used ``as is'':
The FCN-8s architecture is based on a VGG16 \cite{simonyan_very_2015} backbone; 
DeepLabv3+ uses a ResNet-101 \cite{he_deep_2015}; 
The U-Net architecture is identical to the one 
in \cite{ronneberger_u-net_2015}. 

In the experiments with RSS-Net, the number of down-sampling layers in the encoding part has been reduced to be trained with lower resolution inputs. 

The RAMP-CNN architecture dedicated to the RD view has been adapted with two major changes. Firstly, the fusion module has been modified to aggregate and duplicate the feature maps to suit the range-Doppler space. Secondly, the size of the output feature maps has been reduced on the Doppler axis using an additional convolutional layer with $3 \times 3$ filters and a stride factor of 4.
For both RD and RA segmentation tasks, an additional 1D convolutional layer with a softmax operation processes the last feature maps to predict segmentation maps.

\subsection{Training and results}
\label{sec:train_and_res}

\smallskip\noindent\textbf{Training procedures.~}
Methods presented in Section \ref{sec:method_archis} are trained using CARRADA-Train and CARRADA-Val splits and tested on the CARRADA-Test. At each timestamp of a radar sequence, the provided RAD tensor is processed according to the method presented in Section \ref{sec:radar_theory}. 

For each frame, $q$ past frames are also considered for both training and testing phases: The views from $t-q$ to $t$ are stacked into the time-$t$ input (Fig.\,\ref{fig:teaser}).
For the methods that do not explicitly process the time dimension, $q=2$ for a total input sequence length of 3. Time-based methods using 3D convolutions have specific sequence lengths: $q=8$ for RAMP-CNN and $q=4$ for TMVA-Net.

The competing architectures have been trained with the CE loss, except for the RSS-Net, which is trained with a wCE using the formulation in \cite{kaul_rss-net_2020}. Our methods are trained using the combination of loss terms detailed in Section \ref{sec:combi_loss}. We used our formulation of the wCE loss (Section \ref{sec:wce}) with weights computed on  CARRADA-Train.

All the training procedures use the Adam optimiser \cite{kingma_adam_2015} with the recommended parameters ($\beta_1 = 0.9$, $\beta_2 = 0.999$ and $\varepsilon=10^{-8}$). Since each method has its own set of hyper-parameters, further details are provided in Appendix \ref{sec:preproc_procedures}, namely batch sizes, learning rates, learning rate decays, numbers of epochs and corresponding pre-processing steps for each one of them. Training was performed using the PyTorch framework with a single GeForce RTX 2080 Ti graphic card.

\smallskip\noindent\textbf{Quantitative results.~}
The performance for both RD and RA semantic segmentation tasks on CARRADA-Test are shown in Table \ref{table:main_quanti_results}. Our proposed TMVA-Net
achieves the best scores for both mDice and mIoU metrics and for both segmentation tasks.
Moreover, our methods are the only ones to perform both tasks simultaneously. 
TMVA-Net also provides the best trade-off between 
performance and number of parameters for both tasks, as illustrated in Figure \ref{fig:perf_params} with mDice metric (similar plots with mIoU metric are presented in 
Appendix \ref{sec:quanti_results}). Note that the number of parameters reported for each method in Table \ref{table:main_quanti_results}, in Figure \ref{fig:perf_params} and in Appendix \ref{sec:quanti_results} corresponds to a single trained model, while competing methods require two independent models (hence twice more parameters) to perform both RD and RA segmentation tasks.

\begin{table}
\begin{center}
\resizebox{.48\textwidth}{!}{
\begin{tabular}{c l r c c }
\toprule
View & Method & \# Param. & mIoU & mDice \\
\midrule
\multirow{4}{*}{\textbf{RD}} 
& MV-Net (baseline) & 2.4M & 29.0 & 32.8 \\
& MVA-Net (a) & 3.6M & 48.9 & 60.4 \\
& MVA-Net (b) & 4.8M & 52.9 & 64.3\\
& TMVA-Net & 5.6M & \textbf{59.3} & \textbf{71.5}\\
\midrule
\multirow{4}{*}{\textbf{RA}} 
& MV-Net (baseline) & 2.4M & 26.8 & 28.5 \\
& MVA-Net (a) & 3.6M & 28.1 & 31.1 \\
& MVA-Net (b) & 4.8M & 36.7 & 43.9\\
& TMVA-Net & 5.6M & \textbf{40.1} & \textbf{49.3}\\
\bottomrule
\end{tabular}}
\end{center}
\caption{\textbf{Ablation study of the proposed architectures.} Each architecture has been trained using 
the wCE$+$SDice combination loss.
TMVA-Net delivers the best performances under both mIoU and mDice metrics and for both RD and RA views.} 
\label{table:ablation_architectures}
\end{table}

\begin{figure*}[t]
\begin{center}
\includegraphics[width=1\linewidth]{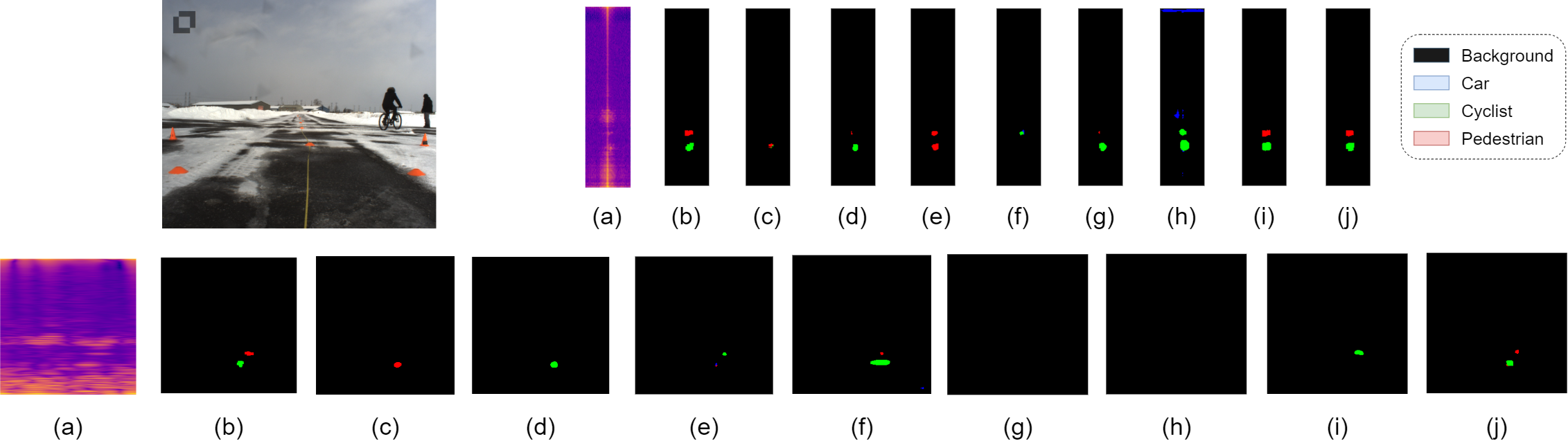}
\end{center}
   \caption{\textbf{Qualitative results on a test scene of CARRADA}. (\textit{Top}) camera image of the scene and results of the RD segmentation; (\textit{Bottom}) Results of the RA Segmentation. (a) Radar view signal, (b) ground-truth mask, (c) FCN8s, (d) U-Net, (e) DeepLabv3+, (f) RSS-Net, (g) RAMP-CNN, (h) MV-Net (our baseline w/ wCE$+$SDice loss), (i) TMVA-Net (ours, w/ wCE$+$SDice loss), (j) TMVA-Net (ours, w/ wCE$+$SDice$+$CoL loss).}
\label{fig:quali_results}
\end{figure*}

\smallskip\noindent\textbf{Ablation Studies.~} 
We report in Table \ref{table:ablation_architectures} the performance of the four architectures we have introduced in Section \ref{sec:archi_ours}. It shows that the additional ASPP modules in MVA-Net(a) boosts the performance relative to MV-Net for both RD and RA segmentation. 
The performance is further improved with MVA-Net(b) by the  additional encoder that extracts features from the AD view and provides relevant information to separate object signatures. 
Finally, TMVA-Net is the most effective regardless of the metric, thanks to its ability to learn spatio-temporal features with 3D convolutions. The temporal dimension indeed helps distinguish between objects and the speckle noise, and to categorise them according to the shape variations.

We assess in Table \ref{table:ablation_losses} the importance of the different losses on the performance of TMVA-Net.
The best combination of two loss terms is wCE$+$SDice for both tasks. The performance is further improved on the RA segmentation task by adding the proposed CoL term, while slightly reduced on RD views. This loss improves the coherence between the tasks by better detecting objects in the range-angle views as discussed in the following section.

\begin{table}
\begin{center}
\resizebox{.48\textwidth}{!}{
\begin{tabular}{lccccc}
\toprule
     & \multicolumn{2}{c}{RD view} & & \multicolumn{2}{c}{RA view} \\
     \cmidrule{2-3} \cmidrule{5-6}
Loss & mIoU & mDice && mIoU & mDice \\
\midrule
CE & 56.1 & 67.8 & &  39.1 & 48.3 \\
SDice & 58.5 & 70.3 & & 37.1 & 44.8\\
wCE & 51.1 & 62.8 && 34.3 & 41.1  \\
CE$+$SDice & 45.2 & 54.0 & & 38.8 & 46.9 \\
wCE$+$SDice & \textbf{59.3} & \textbf{71.5} && \underline{40.1} & \underline{49.3} \\
wCE$+$SDice$+$CoL & \underline{58.7} & \underline{70.9} & & \textbf{41.3} & \textbf{51.0}\\
\bottomrule
\end{tabular}}
\end{center}
\caption{\textbf{Ablation study of the combination of losses.} Each individual or combination of loss(es) is used to train a TMVA-Net model. Our proposed combination (wCE$+$SDice$+$CoL) reaches the best mIoU and mDice for the RA view and the second best scores for the RD view.}
\label{table:ablation_losses}
\end{table}

\smallskip\noindent\textbf{Qualitative results.~} We show in Figure \ref{fig:quali_results} qualitative results of each method on a scene from CARRADA-Test. The results of TMVA-Net (i-j) display well segmented RD views in terms of localisation and classification. Only TMVA-Net with CoL (j) is able to localise and classify both objects in the RD and RA views. 
The enforcement of the coherence of predictions across views succeeds in correctly classifying the same objects in the two views. This is not the case for TMVA-Net without CoL, as illustrated in the example (i), where the model predicts a cyclist instead of a pedestrian in the RA view. Moreover, the coherence loss also helps to discover new objects: In (i), TMVA-Net predicts a single object in the RA view, while in (j),  it localises and classifies both objects well with the help of CoL.

Qualitative results on additional urban scenes, which are not part of CARRADA, are provided in Appendix \ref{sec:quali_results}. They show that our methods unlike others can generalise well on RD and RA views. 
Indeed, TMVA-Net succeeds in learning object signatures on the CARRADA dataset and recognizing them in a different environment.

\section{Conclusions and perspectives}
\label{sec:conclusions}

In this work, we propose lightweight architectures for multi-view radar semantic segmentation and a combination of loss terms to train them. Our methods localise and delineate objects in the radar scene while simultaneously determining their relative velocity. We show that both the information from the RAD radar tensor and from its temporal evolution are important to 
conduct these tasks.
Our methods significantly outperform competing architectures specialised either in natural image semantic segmentation or in radar scene understanding. 
Preliminary experiments also show qualitatively that they generalize better to new complex urban scenes. 

We hope that our work will encourage the community to release larger datasets to emphasize the importance of the radar sensor 
and to explore new architectures for radar semantic segmentation. 

Our future investigations will focus on improving the segmentation of cyclists and pedestrians, which remain difficult to distinguish. 
Exploiting radar properties could be interesting to improve both RAD tensor aggregation and class-specific data augmentation methods for the benefit of our learning algorithms.

\paragraph{Acknowledgements} We thank Veronica Elizabeth Vargas Salas for her valuable help with temporal radar data.

\newpage

{\small
\bibliographystyle{ieee_fullname}
\bibliography{iccv2021_biblio}
}

\newpage
\appendix
\section{RAD tensor visualisation}
\label{sec:rad_tensor_vis}

\begin{figure*}
\begin{center}
\includegraphics[width=1\linewidth]{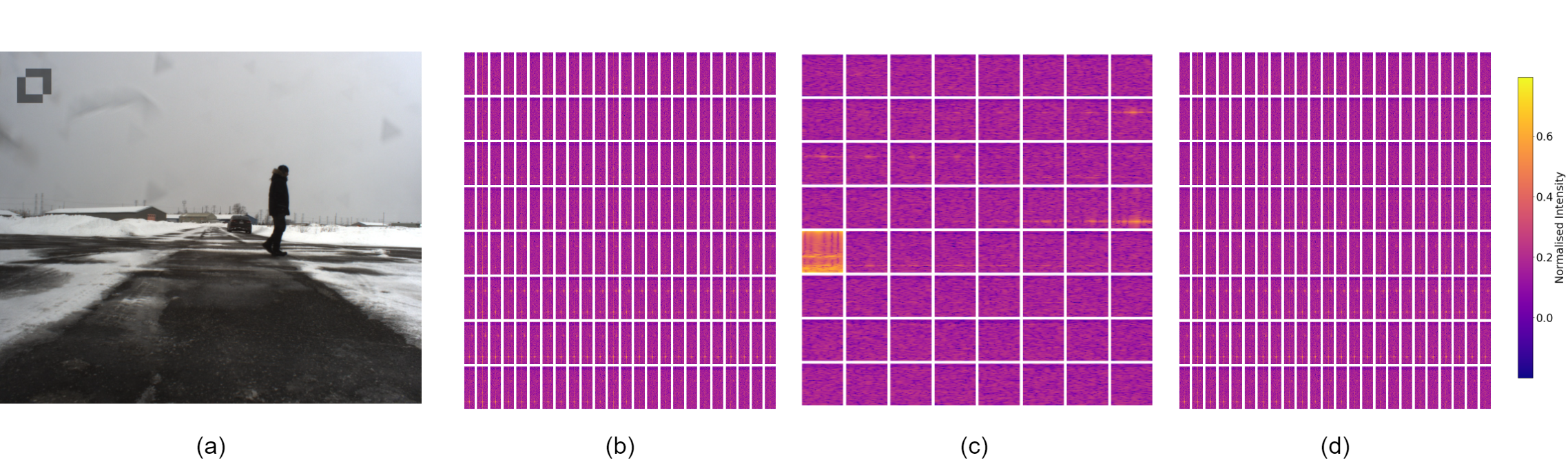}
\end{center}
   \caption{\textbf{Visualisation of the Range-Angle-Doppler (RAD) tensor.} (a) Camera image of the scene. The corresponding RAD tensor is visualised by slices of (b) range-Doppler, (c) range-angle or (d) angle-Doppler views w.r.t. their discretized third axis.}
\label{fig:rad_tensor}
\end{figure*}

An illustration of the RAD tensor is proposed in Figure~\ref{fig:rad_tensor}. Each slice of 2D views corresponds to a discretized bin of the third axis. In Figure \ref{fig:rad_tensor}(b) for instance, the $256$ range-Doppler slices correspond to the view of each discretized value of the angle axis. One can observe redundant signal information and a significant level of noise for each group of 2D-view slices.

\section{Detailed multi-view architectures}
\label{sec:archi_details}

We provide in Tables\,\ref{table:mvnet_archi} and \ref{table:tmvanet_archi} the architecture details of the proposed multi-view network (MV-Net) and temporal multi-view network with ASPP modules (TMVA-Net) respectively. 
For each layer, the parameters of the operations used are specified in the following manner:
\begin{itemize}
    \item $n$-dim convolution: Conv$n$D\,(input\_channels, output\_channels, kernel\_size, stride, padding, dilation\_rate);
    \item \mbox{$n$-dim up-convolution: ConvTranspose$n$D\,(input\_chan-}\\nels, output\_channels, kernel\_size, stride, padding, dilation\_rate);
    \item maximum pooling: MaxPool2D\,(kernel\_size, stride);
    \item atrous spatial pyramid pooling: ASPP\,(input\_channels, output\_channels);
    \item $n$-D batch normalisation: BN$n$D\,(input\_channels);
    \item Leaky ReLU activation:  LeakyReLU\,(negative\_slope);
\end{itemize}
Where $n\!\in\!\{1,2,3\}$ is the dimension of the associated operation.

The ASPP module \cite{chen_deeplab_2017} has the same architecture as the one introduced by Kaul \etal \cite{kaul_rss-net_2020} for range-angle semantic segmentation. We note that the `output\_channels' parameter for the ASPP module corresponds to the number of output channels for each parallel convolution. We also note that the `stride' parameter can be either a scalar or a tuple of scalars depending on the axis on which it is applied.

\begin{table*}
\begin{center}
\scriptsize
\renewcommand{\arraystretch}{1.8}
\begin{tabular}{c c c c l}
\toprule
 & Layer & Inputs & \makecell[c]{Output resolution \\ ($C \times H \times W$)} & Operations \\
\midrule
\multirow{5}{*}{\textbf{RD Encoder}} 
& rd\_layer1 & RD view & $128 \times 256 \times 64$ & \makecell[l]{Conv2D(3, 128, $3 \times 3$, 1, 1, 1) $+$ BN2D(128) $+$ LeakyReLU(0.01) \\ Conv2D(128, 128, $3 \times 3$, 1, 1, 1) $+$ BN2D(128) $+$ LeakyReLU(0.01)}\\

& rd\_layer2 & rd\_layer1 & $128 \times 128 \times 64$ & MaxPool2D(2, (2, 1)) \\

& rd\_layer3 & rd\_layer2 & $128 \times 128 \times 64$ & \makecell[l]{Conv2D(128, 128, $3 \times 3$, 1, 1, 1) $+$ BN2D(128) $+$ LeakyReLU(0.01) \\ Conv2D(128, 128, $3 \times 3$, 1, 1, 1) $+$ BN2D(128) $+$ LeakyReLU(0.01)}\\

& rd\_layer4 & rd\_layer3 & $128 \times 64 \times 64$ & MaxPool2D(2, (2, 1)) \\

& rd\_layer5 & rd\_layer4 & $128 \times 64 \times 64$ & Conv1D(128, 128, $1 \times 1$, 1, 0, 1)\\

\midrule
\multirow{5}{*}{\textbf{RA Encoder}} 
& ra\_layer1  & RA view & $128 \times 256 \times 256$ & \makecell[l]{Conv2D(3, 128, $3 \times 3$, 1, 1, 1) $+$ BN2D(128) $+$ LeakyReLU(0.01) \\ Conv2D(128, 128, $3 \times 3$, 1, 1, 1) $+$ BN2D(128) $+$ LeakyReLU(0.01)} \\

& ra\_layer2 & ra\_layer1 & $128 \times 128 \times 128$ & MaxPool2D(2, 2)\\

& ra\_layer3 & ra\_layer2 & $128 \times 128 \times 128$ & \makecell[l]{Conv2D(128, 128, $3 \times 3$, 1, 1, 1) $+$ BN2D(128) $+$ LeakyReLU(0.01) \\ Conv2D(128, 128, $3 \times 3$, 1, 1, 1) $+$ BN2D(128) $+$ LeakyReLU(0.01)} \\

& ra\_layer4 & ra\_layer3 & $128 \times 64 \times 64$ & MaxPool2D(2, 2)\\

& ra\_layer5 & ra\_layer4  & $128 \times 64 \times 64$ & Conv1D(128, 128, $1 \times 1$, 1, 0, 1)\\

\midrule
\textbf{Latent space} & layer6 & rd\_layer5, ra\_layer5 & $256 \times 64 \times 64$ &	concatenate(rd\_layer5, ra\_layer5) \\

\midrule
\multirow{6}{*}{\textbf{RD Decoder}} 
& rd\_layer7 & layer6 & $128 \times 64 \times 64$ & Conv1D(256, 128, $1 \times 1$, 1, 0, 1) \\

& rd\_layer8 & rd\_layer7 & $128 \times 128 \times 64$ & ConvTranspose2D(128, 128, $2 \times 1$, (2, 1), 1, 1) \\

& rd\_layer9 & rd\_layer8 & $128 \times 128 \times 64$ & \makecell[l]{Conv2D(128, 128, $3 \times 3$, 1, 1, 1) $+$ BN2D(128) $+$ LeakyReLU(0.01) \\ Conv2D(128, 128, $3 \times 3$, 1, 1, 1) $+$ BN2D(128) $+$ LeakyReLU(0.01)} \\

& rd\_layer10 & rd\_layer9 & $128 \times 256 \times 64$	& ConvTranspose2D(128, 128, $2 \times 1$, (2, 1), 1, 1) \\

& rd\_layer11 & rd\_layer10 & $128 \times 256 \times 64$ & \makecell[l]{Conv2D(128, 128, $3 \times 3$, 1, 1, 1) $+$ BN2D(128) $+$ LeakyReLU(0.01) \\ Conv2D(128, 128, $3 \times 3$, 1, 1, 1) $+$ BN2D(128) $+$ LeakyReLU(0.01)} \\

& rd\_layer12 & rd\_layer11 & $K \times 256 \times 64$ & Conv1D(128, $K$, $1 \times 1$, 1, 0, 1) \\

\midrule
\multirow{6}{*}{\textbf{RA Decoder}} 
& ra\_layer7 & layer6 &	$128 \times 64 \times 64$ & Conv1D(256, 128, $1 \times 1$, 1, 0, 1) \\

& ra\_layer8 & ra\_layer7 & $128 \times 128 \times 128$ & ConvTranspose2D(128, 128, $2 \times 2$, 2, 1, 1) \\

& ra\_layer9 & ra\_layer8 & $128 \times 128 \times 128$ & \makecell[l]{Conv2D(128, 128, $3 \times 3$, 1, 1, 1) $+$ BN2D(128) $+$ LeakyReLU(0.01) \\ Conv2D(128, 128, $3 \times 3$, 1, 1, 1) $+$ BN2D(128) $+$ LeakyReLU(0.01)} \\

& ra\_layer10 & ra\_layer9 & $128 \times 256 \times 256$ & ConvTranspose2D(128, 128, $2 \times 2$, 2, 1, 1) \\

& ra\_layer11 & ra\_layer10 & $128 \times 256 \times 256$ & \makecell[l]{Conv2D(128, 128, $3 \times 3$, 1, 1, 1) $+$ BN2D(128) $+$ LeakyReLU(0.01) \\ Conv2D(128, 128, $3 \times 3$, 1, 1, 1) $+$ BN2D(128) $+$ LeakyReLU(0.01)} \\

& ra\_layer12 & ra\_layer11 & $K \times 256 \times 256$ & Conv1D(128, $K$, $1 \times 1$, 1, 0, 1) \\

\bottomrule
\end{tabular}
\end{center}
\caption{\textbf{Multi-view network (MV-Net) architecture.} This table lists all the layers contained in the model taking as input multi-view radar representations (RD and RA views) to predict segmentation maps for each multi-view output. 
Details about the parameters of each operation are provided in Section \ref{sec:archi_details}.
We note $K$ the number of classes. 
The number of input channels in the first layer corresponds to the consecutive frames of each view stacked in temporal dimension, here $q=2$ and thus the number of channels is $3$.}
\label{table:mvnet_archi}
\end{table*}

\begin{table*}
\begin{center}
\scriptsize
\renewcommand{\arraystretch}{1.8}
\begin{tabular}{c c c c l}
\toprule
 & Layer & Inputs & \makecell[c]{Output resolution \\ ($C \times H \times W$)} & Operations \\
\midrule
\multirow{7}{*}{\textbf{RD Encoder}} 
& rd\_layer1 & RD view & $128 \times 256 \times 64$ & \makecell[l]{Conv3D(1, 128, $3 \times 3 \times 3$, 1, (0, 1, 1), 1) $+$ BN3D(128) $+$ LeakyReLU(0.01) \\ Conv3D(128, 128, $3 \times 3 \times 3$, 1, (0, 1, 1), 1) $+$ BN3D(128) $+$ LeakyReLU(0.01)} \\

& rd\_layer2 & rd\_layer1 & $128 \times 128 \times 64$ & MaxPool2D(2, (2, 1)) \\

& rd\_layer3 & rd\_layer2 & $128 \times 128 \times 64$ & \makecell[l]{Conv2D(128, 128, $3 \times 3$, 1, 1, 1) $+$ BN2D(128) $+$ LeakyReLU(0.01) \\ Conv2D(128, 128, $3 \times 3$, 1, 1, 1) $+$ BN2D(128) $+$ LeakyReLU(0.01)}\\

& rd\_layer4 & rd\_layer3 & $128 \times 64 \times 64$ & MaxPool2D(2, (2, 1)) \\

& rd\_layer5 & rd\_layer4 & $128 \times 64 \times 64$ & Conv1D(128, 128, $1 \times 1$, 1, 0, 1)\\

& rd\_layer6 & rd\_layer5 & $640 \times 64 \times 64$ & ASPP(128, 128) \\

& rd\_layer7 & rd\_layer6 & $128 \times 64 \times 64$ & Conv1D(640, 128,  $1 \times 1$, 1, 0, 1) \\

\midrule
\multirow{7}{*}{\textbf{AD Encoder}} 
& ad\_layer1 & AD view & $128 \times 256 \times 64$ & \makecell[l]{Conv3D(1, 128, $3 \times 3 \times 3$, 1, (0, 1, 1), 1) $+$ BN3D(128) $+$ LeakyReLU(0.01) \\ Conv3D(128, 128, $3 \times 3 \times 3$, 1, (0, 1, 1), 1) $+$ BN3D(128) $+$ LeakyReLU(0.01)} \\

& ad\_layer2 & ad\_layer1 & $128 \times 128 \times 64$ & MaxPool2D(2, (2, 1)) \\

& ad\_layer3 & ad\_layer2 & $128 \times 128 \times 64$ & \makecell[l]{Conv2D(128, 128, $3 \times 3$, 1, 1, 1) $+$ BN2D(128) $+$ LeakyReLU(0.01) \\ Conv2D(128, 128, $3 \times 3$, 1, 1, 1) $+$ BN2D(128) $+$ LeakyReLU(0.01)}\\

& ad\_layer4 & ad\_layer3 & $128 \times 64 \times 64$ & MaxPool2D(2, (2, 1)) \\

& ad\_layer5 & ad\_layer4 & $128 \times 64 \times 64$ & Conv1D(128, 128, $1 \times 1$, 1, 0, 1)\\

& ad\_layer6 & ad\_layer5 & $640 \times 64 \times 64$ & ASPP(128, 128) \\

& ad\_layer7 & ad\_layer6 & $128 \times 64 \times 64$ & Conv1D(640, 128,  $1 \times 1$, 1, 0, 1) \\

\midrule
\multirow{7}{*}{\textbf{RA Encoder}} 

& ra\_layer1 & RA view & $128 \times 256 \times 256$ & \makecell[l]{Conv3D(1, 128, $3 \times 3 \times 3$, 1, (0, 1, 1), 1) $+$ BN3D(128) $+$ LeakyReLU(0.01) \\ Conv3D(128, 128, $3 \times 3 \times 3$, 1, (0, 1, 1), 1) $+$ BN3D(128) $+$ LeakyReLU(0.01)} \\

& ra\_layer2 & ra\_layer1 & $128 \times 128 \times 128$ & MaxPool2D(2, 2)\\

& ra\_layer3 & ra\_layer2 & $128 \times 128 \times 128$ & \makecell[l]{Conv2D(128, 128, $3 \times 3$, 1, 1, 1) $+$ BN2D(128) $+$ LeakyReLU(0.01) \\ Conv2D(128, 128, $3 \times 3$, 1, 1, 1) $+$ BN2D(128) $+$ LeakyReLU(0.01)} \\

& ra\_layer4 & ra\_layer3 & $128 \times 64 \times 64$ & MaxPool2D(2, 2)\\

& ra\_layer5 & ra\_layer4  & $128 \times 64 \times 64$ & Conv1D(128, 128, $1 \times 1$, 1, 0, 1)\\

& ra\_layer6 & ra\_layer5 & $640 \times 64 \times 64$ & ASPP(128, 128) \\

& ra\_layer7 & ra\_layer6 & $128 \times 64 \times 64$ & Conv1D(640, 128,  $1 \times 1$, 1, 0, 1) \\

\midrule
\textbf{Latent space} & layer8 & rd\_layer5, ra\_layer5, ad\_layer5 & $384 \times 64 \times 64$ & concatenate(rd\_layer5, ra\_layer5, ad\_layer5) \\

\midrule
\multirow{7}{*}{\textbf{RD Decoder}} 
& rd\_layer9 & layer8 & $128 \times 64 \times 64$ & Conv1D(384, 128, $1 \times 1$, 1, 0, 1) \\

& rd\_layer10 & rd\_layer7, rd\_layer9, ad\_layer7 & $384 \times 64 \times 64$ & concatenate(rd\_layer7, rd\_layer9, ad\_layer7) \\

& rd\_layer11 & rd\_layer10 & $128 \times 128 \times 64$ & ConvTranspose2D(384, 128, $2 \times 1$, (2, 1), 1, 1) \\

& rd\_layer12 & rd\_layer11 & $128 \times 128 \times 64$ & \makecell[l]{Conv2D(128, 128, $3 \times 3$, 1, 1, 1) $+$ BN2D(128) $+$ LeakyReLU(0.01) \\ Conv2D(128, 128, $3 \times 3$, 1, 1, 1) $+$ BN2D(128) $+$ LeakyReLU(0.01)} \\

& rd\_layer13 & rd\_layer12 & $128 \times 256 \times 64$	& ConvTranspose2D(128, 128, $2 \times 1$, (2, 1), 1, 1) \\

& rd\_layer14 & rd\_layer13 & $128 \times 256 \times 64$ & \makecell[l]{Conv2D(128, 128, $3 \times 3$, 1, 1, 1) $+$ BN2D(128) $+$ LeakyReLU(0.01) \\ Conv2D(128, 128, $3 \times 3$, 1, 1, 1) $+$ BN2D(128) $+$ LeakyReLU(0.01)} \\

& rd\_layer15 & rd\_layer14 & $K \times 256 \times 64$ & Conv1D(128, $K$, $1 \times 1$, 1, 0, 1) \\

\midrule
\multirow{7}{*}{\textbf{RA Decoder}} 
& ra\_layer9 & layer8 & $128 \times 64 \times 64$ & Conv1D(384, 128, $1 \times 1$, 1, 0, 1) \\

& ra\_layer10 & ra\_layer7, ra\_layer9, ad\_layer7 & $384 \times 64 \times 64$ & concatenate(ra\_layer7, ra\_layer9, ad\_layer7) \\

& ra\_layer11 & ra\_layer10 & $384 \times 128 \times 128$ & ConvTranspose2D(128, 128, $2 \times 2$, 2, 1, 1) \\

& ra\_layer12 & ra\_layer11 & $128 \times 128 \times 128$ & \makecell[l]{Conv2D(128, 128, $3 \times 3$, 1, 1, 1) $+$ BN2D(128) $+$ LeakyReLU(0.01) \\ Conv2D(128, 128, $3 \times 3$, 1, 1, 1) $+$ BN2D(128) $+$ LeakyReLU(0.01)} \\

& ra\_layer13 & ra\_layer12 & $128 \times 256 \times 256$ & ConvTranspose2D(128, 128, $2 \times 2$, 2, 1, 1) \\

& ra\_layer14 & ra\_layer13 & $128 \times 256 \times 256$ & \makecell[l]{Conv2D(128, 128, $3 \times 3$, 1, 1, 1) $+$ BN2D(128) $+$ LeakyReLU(0.01) \\ Conv2D(128, 128, $3 \times 3$, 1, 1, 1) $+$ BN2D(128) $+$ LeakyReLU(0.01)} \\

& ra\_layer15 & ra\_layer14 & $K \times 256 \times 256$ & Conv1D(128, $K$, $1 \times 1$, 1, 0, 1) \\

\bottomrule
\end{tabular}
\end{center}
\caption{\textbf{Temporal multi-view network with ASPP modules (TMVA-Net) architecture.} This table lists all the layers contained in the model taking as input multi-view radar representations (RD and RA views) to predict segmentation maps for each multi-view output.
Details about the parameters of each operation are provided in Sec. \ref{sec:archi_details}. 
We note $K$ the number of classes. 
The number of input channels in the first layer is fixed to $1$ because the consecutive frames are considered as a sequence, here $q=4$ and thus the number of channels is $5$.}

\label{table:tmvanet_archi}
\end{table*}


\section{Coherence loss}
\label{sec:col_figure}

\begin{figure}[t]
\begin{center}
\includegraphics[width=1\linewidth]{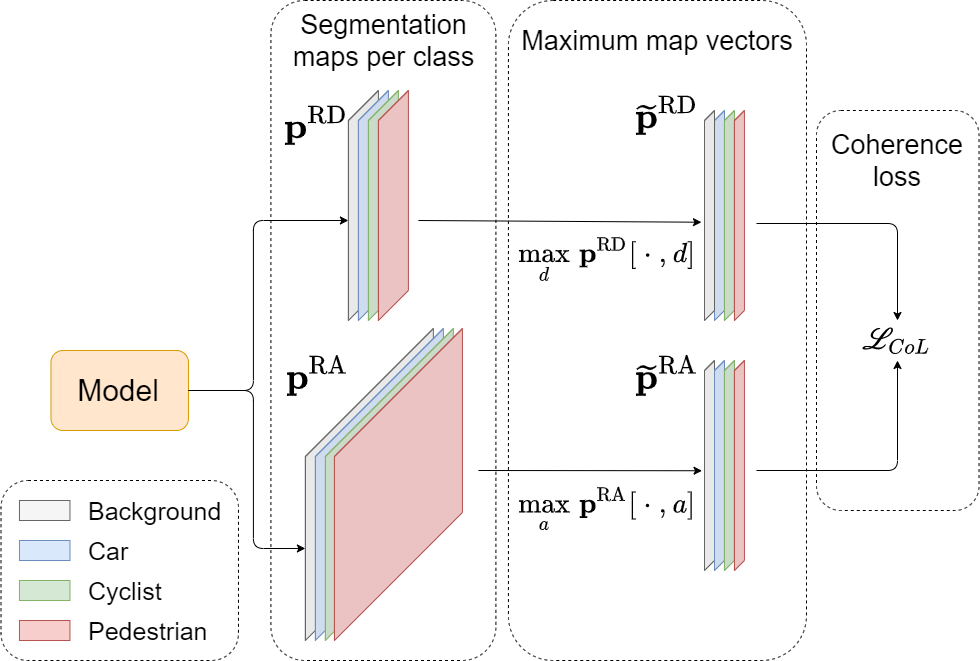}
\end{center}
   \caption{\textbf{Computation of the coherence loss.} The segmentation maps $\vect{p}^{\text{RD}}$ and $\vect{p}^{\text{RA}}$ of the two views are aggregated by max pooling along the axis
   that they do not share (either the Doppler or the angle). The coherence loss is the mean squared error (MSE) between the two resulting vectors $\tilde{\vect{p}}^{\text{RD}}$ and $\tilde{\vect{p}}^{\text{RA}}$.}
\label{fig:coherence_loss}
\end{figure}

The purpose of the coherence loss (CoL) is to preserve a consistency between the predictions of the model for the different views of the same scene.
The procedure used to construct this loss is illustrated in Figure \ref{fig:coherence_loss}.

\section{Pre-processing and training procedures}
\label{sec:preproc_procedures}

The experiments in the main paper have been conducted using the parameters detailed in Table \ref{table:details_training}. 
An exponential decay with $\gamma = 0.9$ has been applied to each learning rate with an epoch step specific to each model (see Table \ref{table:details_training}).
The competing methods have been trained using the Cross Entropy (CE) loss, except for the RSS-Net, which is trained with a
weighted Cross Entropy (wCE) using the formulation in \cite{kaul_rss-net_2020}. Our methods have been trained with the proposed combination of losses using the following parameters set up empirically: $\lambda_{\text{wCE}}= 1$, $\lambda_{\text{SDice}} = 10$ and $\lambda_{\text{CoL}}= 5$.

The architectures with which we compare our work have been designed to process inputs of size $256 \times 256$. Since the size of the range-Doppler view is $256 \times 64$ in the CARRADA dataset \cite{ouaknine_carrada_2020}, it is resized in the Doppler dimension to train these competing models. 
On the other hand, our proposed architectures are composed of down-sampling layers adapted to the size of the Doppler dimension, thus they do not require this pre-processing step.
The range-angle view has a size of $256 \times 256$ and does not require a resizing in both cases. 
For all methods, we used vertical and horizontal flip as data augmentation to reduce over-fitting.

Each view is normalised between 0 and 1 using local batch statistics for the competing methods. Our normalisation strategy consists in using the global statistics of the entire CARRADA dataset to normalise the input views.

\begin{SCtable*}
\resizebox{.47\textwidth}{!}{
\begin{tabular}{c l r c c c c c }
\toprule
View & Method & Param. \# & $q$ & Batch size & LR & LR step & Epoch \#  \\
\midrule
\multirow{7}{*}{\textbf{RD}} 
& FCN-8s \cite{long_fully_2015} & 134.3~~ & 0 & 20 & $ 10^{-4}$ & 10 & 100 \\
& U-Net \cite{ronneberger_u-net_2015} & 17.3~~ & 3 & 6 & $ 10^{-4}$ & 20 & 150 \\
& DeepLabv3+ \cite{chen_encoder-decoder_2018} & 59.3~~ & 3 & 20 & $ 10^{-4}$ & 20 & 150 \\
& RSS-Net & 10.1~~ & 3 & 6 & $ 10^{-3}$ & 10 & 100 \\
& RAMP-CNN & 106.4~~ & 9 & 2 & $ 10^{-5}$ & 20 & 150 \\
& MV-Net (ours-baseline) & 2.4* & 3 & 13 & $ 10^{-4}$ & 20 & 300 \\
& TMVA-Net (ours) & 5.6* & 5 & 6 & $ 10^{-4}$ & 20 & 300 \\
\midrule
\multirow{7}{*}{\textbf{RA}}
& FCN-8s \cite{long_fully_2015} & 134.3~~ & 0 & 10 & $ 10^{-4}$ & 10 & 100 \\
& U-Net \cite{ronneberger_u-net_2015} & 17.3~~ & 3 & 6 & $ 10^{-4}$ & 20 & 150 \\
& DeepLabv3+ \cite{chen_encoder-decoder_2018} & 59.3~~ & 3 & 20 & $ 10^{-4}$ & 20 & 150 \\
& RSS-Net & 10.1~~ & 3 & 6 & $ 10^{-4}$ & 10 & 100 \\
& RAMP-CNN & 106.4~~ & 9 & 2 & $ 10^{-5}$ & 20 & 150 \\
& MV-Net (ours-baseline) & 2.4* & 3 & 13 & $ 10^{-4}$ & 20 & 300 \\
& TMVA-Net (ours) & 5.6* & 5 & 6 & $ 10^{-4}$ & 20 & 300 \\
\bottomrule
\end{tabular}}
\caption{\textbf{Hyper-parameters used for training.} The number of trainable parameters (in millions) for each method corresponds to a single view-segmentation model; Two such models, one for each view, are required for all methods but ours. In contrast, the number of parameters reported for our methods (`*') corresponds to a single model that segments both RD and RA views. RSS-Net and RAMP-CNN have been modified to be trained on both tasks (see Sec.\,4.2 of the main article). The input of a model consists in $q+1$ successive RAD frames, where $q$ is the number of considered past frames, if any. The learning rate (`LR') step is in epochs.}
\label{table:details_training}
\end{SCtable*}

\section{Quantitative results}
\label{sec:quanti_results}
\begin{figure}[t]
\begin{center}
   \includegraphics[width=1\linewidth]{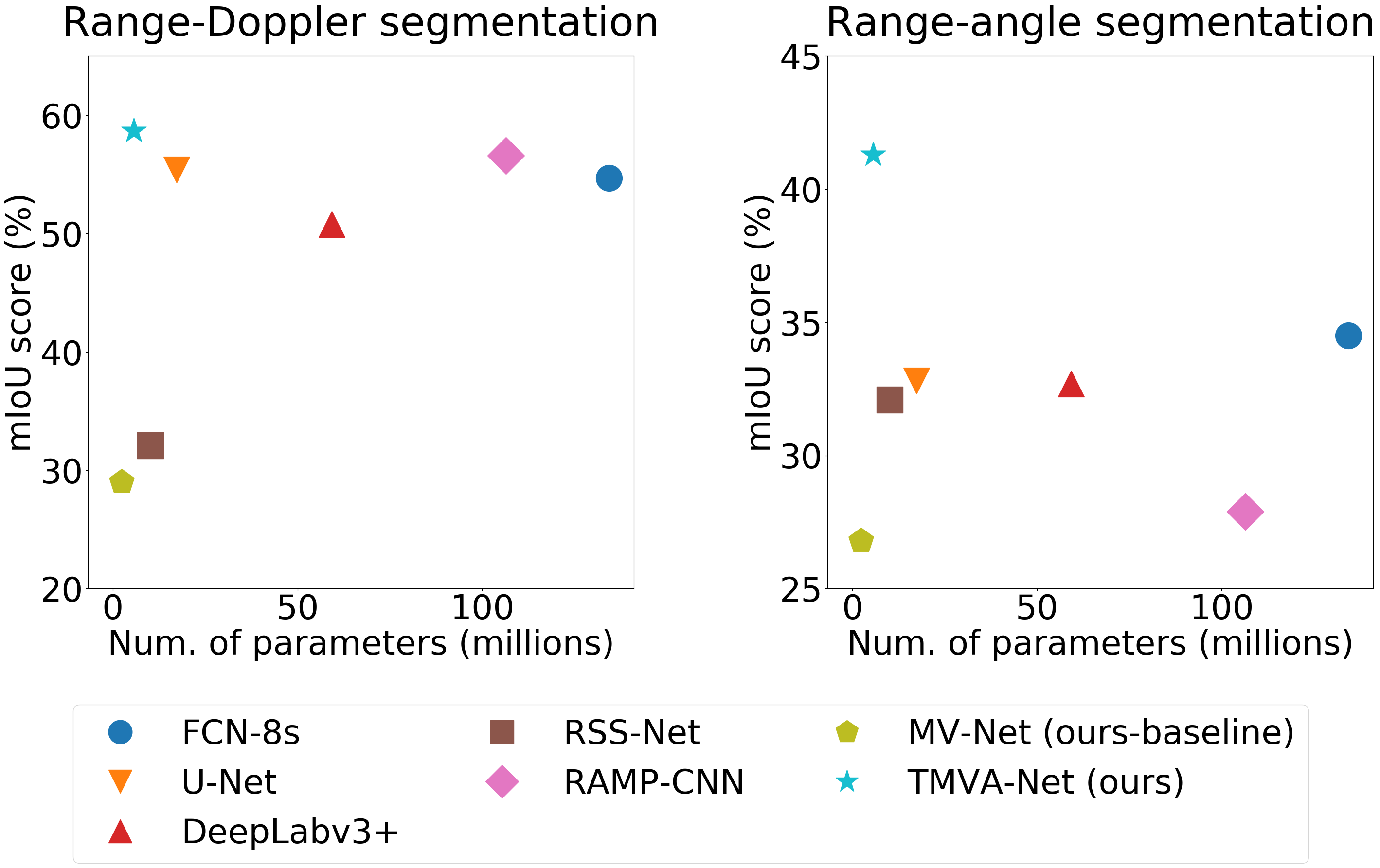}
\end{center}
   \caption{\textbf{Performance-\textit{vs}.-complexity plots for all methods in RD and RA tasks.} Performance is assessed by mIoU (\%) and complexity by the number of parameters (in millions) \textit{for a single task}. 
   Top-left models correspond to the best performing and the lightest. 
    Only our models, MV-Net and TMVA-Net, are able to segment both views simultaneously. For all the other methods, two distinct models must be trained to address both tasks, which doubles the number of actual parameters.}
\label{fig:perf_params_iou}
\end{figure}

Our proposed TMVA-Net architecture provides the best trade-off between performance and number of parameters for both range-Doppler and range-angle semantic segmentation tasks, as illustrated in Figure \ref{fig:perf_params_iou} with mIoU metric.

\section{Qualitative results}
\label{sec:quali_results}

\begin{figure*}
\begin{center}
\includegraphics[width=1\linewidth]{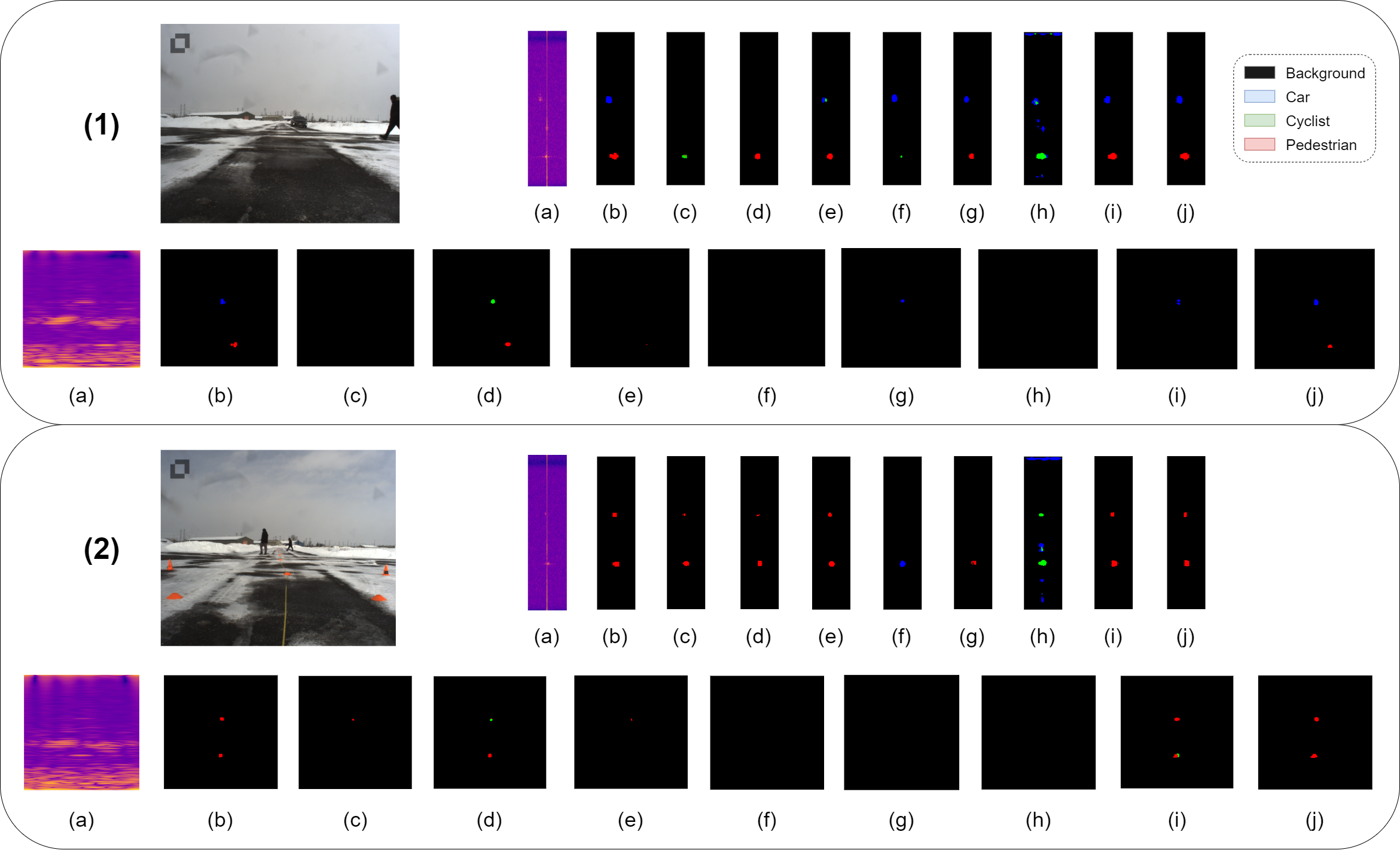}
\end{center}
   \caption{\textbf{Qualitative results on two test scenes of CARRADA-Test}. (1) and (2) are two independent examples. (\textit{Top}) camera image of the scene and results of the RD segmentation; (\textit{Bottom}) Results of the RA Segmentation. (a) Radar view signal, (b) ground-truth mask, (c) FCN8s, (d) U-Net, (e) DeepLabv3+, (f) RSS-Net, (g) RAMP-CNN, (h) MV-Net (our baseline w/ wCE$+$SDice loss), (i) TMVA-Net (ours, w/ wCE$+$SDice loss), (j) TMVA-Net (ours, w/ wCE$+$SDice$+$CoL loss).}
\label{fig:quali_results_supp_carrada}
\end{figure*}

Additional qualitative results are shown in Figure \ref{fig:quali_results_supp_carrada} for each method on scenes (1-2) from the CARRADA-Test. For the scene (1), RAMP-CNN (g) and TMVA-Net (i-j) display well segmented RD views. However, only TMVA-Net with CoL (j) is able to localise and classify both objects in the RD and RA views of the first example. In scene (2), four methods (d-e-i-j) are able to well localise objects in the RD view. Once again, only TVMA-Net with CoL (j) is able to well segment objects in both RD and RA views while our method without CoL (i) predicts pedestrian and cyclist categories for pixels of the same object.

\begin{figure*}
\begin{center}
\includegraphics[width=1\linewidth]{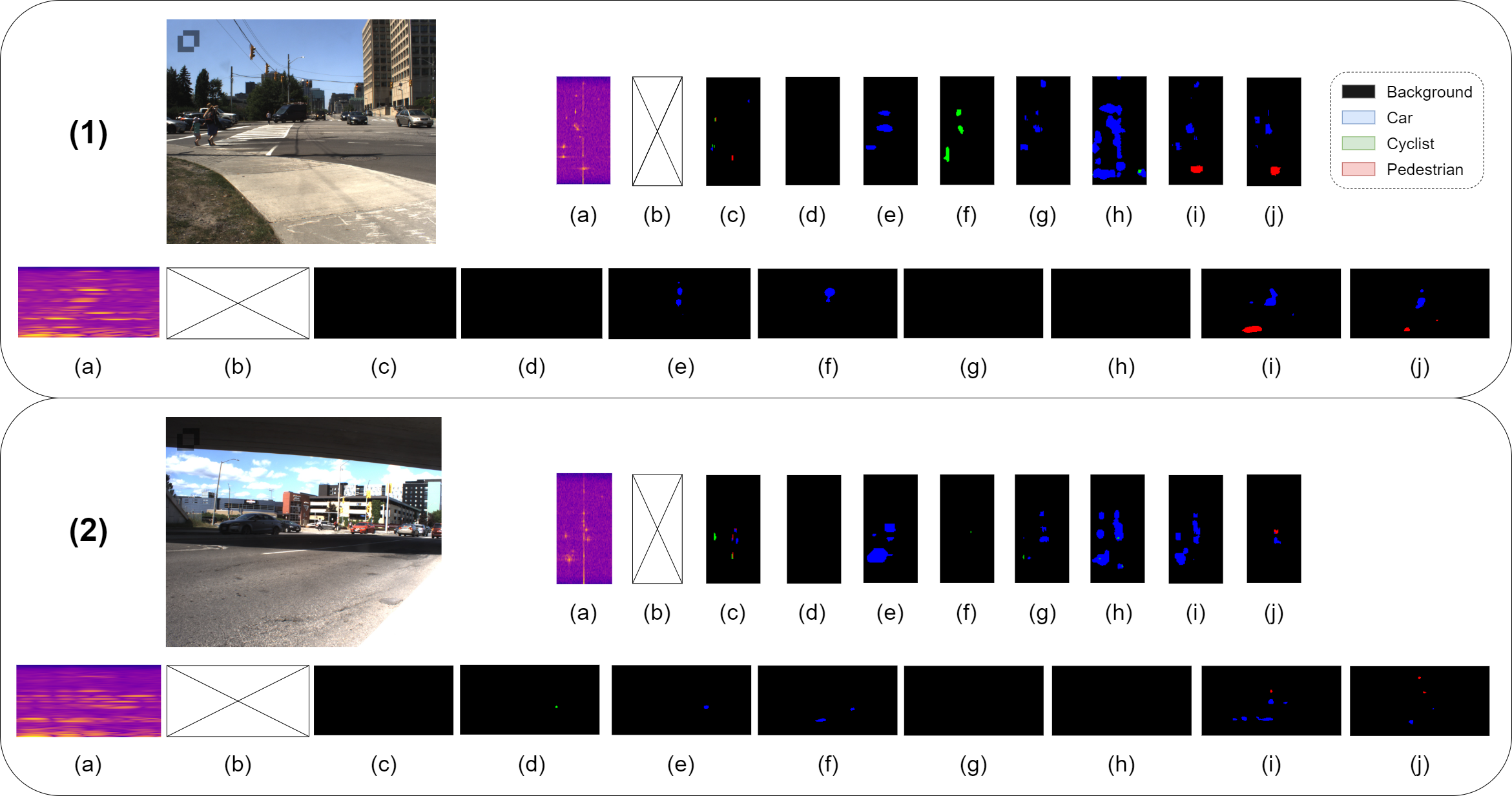}
\end{center}
   \caption{\textbf{Qualitative results on two test scenes of in-house sequences}. (1) and (2) are two independent examples. (\textit{Top}) camera image of the scene and results of the RD segmentation; (\textit{Bottom}) Results of the RA Segmentation. (a) Radar view signal, (b) ground-truth mask, (c) FCN8s, (d) U-Net, (e) DeepLabv3+, (f) RSS-Net, (g) RAMP-CNN, (h) MV-Net (our baseline w/ wCE$+$SDice loss), (i) TMVA-Net (ours, w/ wCE$+$SDice loss), (j) TMVA-Net (ours, w/ wCE$+$SDice$+$CoL loss).}
\label{fig:quali_results_supp_sc}
\end{figure*}

Figure\,\ref{fig:quali_results_supp_sc} shows qualitative results for each method trained on CARRADA-Train and CARRADA-Val, and tested on in-house sequences of complex urban scenes (1-2) with a different range resolution. 
The qualitative examples and results have been cropped with respect to the minimum and maximum range of the dataset used for training. The ground-truth masks in columns (1-b) and (2-b) are empty because the radar views are not annotated.
In scene (1), only TMVA-Net models (i-j) are able to localise and classify the signals related to the pedestrians and cars in both the RD and the RA views. 
In scene (2), only TMVA-Net (i-j) methods succeed to localise and classify cars and pedestrians in the RA view. We note that TMVA-Net without CoL (i) detects more car signals while TMVA-Net with CoL (j) is the only method capable of distinguishing pedestrian signatures on both RD and RA views.

These two examples in complex urban scenes suggest that our method has successfully learnt object signatures in the CARRADA dataset and is able to generalise well.

\end{document}